\documentclass[11pt]{article}
\usepackage[preprint]{acl}
\usepackage{times}
\usepackage{latexsym}
\usepackage[T1]{fontenc}
\usepackage[utf8]{inputenc}
\usepackage{microtype}
\usepackage{inconsolata}
\usepackage{graphicx}
\usepackage{amsmath}
\usepackage{amssymb}
\usepackage{booktabs}
\usepackage{placeins}
\usepackage{float}
\usepackage{listings}
\usepackage{xcolor}
\title{Automated Mediator for Human Negotiation: Pre-Mediation via a Structured LLM Pipeline}
\author{
 Jamie Bergen, Sarit Kraus \\
}
\begin{document}
\maketitle
\begin{abstract}
Pre-mediation, the preparatory phase preceding direct human negotiation, plays a critical role in achieving mutually beneficial agreements, yet is often omitted due to cost, time constraints, and limited access to trained mediators. We introduce an automated mediator for human negotiation, implemented as a structured pipeline of LLM modules, that supports pre-mediation in integrative negotiation settings. The pipeline decomposes preparation into specialized LLM modules for dialogue, preference prediction, response-level critique, and structured summarization, separating inference, generation, and evaluation to address limitations of monolithic single-prompt approaches. Although we use the term ``agent'' to refer to each module (following common LLM-systems terminology), the components are not autonomous and do not interact peer-to-peer; outputs are passed forward in a fixed sequence.
We evaluate the system in two controlled human-subject experiments comparing AI-based pre-mediation with professional human mediators in a multi-issue negotiation scenario. Results show that, on short-term self-reported measures, the automated mediator achieves preparation outcomes broadly comparable to human mediators---including trust in the mediator and confidence in reaching mutually beneficial agreements---while achieving substantially lower error on the preference-inference task under our scenario and prompts (36\% lower RMSE). A second study demonstrates that targeted prompt refinements successfully reduce excessive affirmation patterns from 36.6\% to 16.8\%, matching human mediator baselines.
Overall, our findings suggest that structured LLM pipelines can provide scalable, low-effort pre-mediation support broadly comparable to human mediators on short-term, self-reported preparation outcomes. The pipeline's single-party design is a deliberate architectural feature: it mirrors how human mediators run pre-mediation today and enables parallel deployment across all parties to a dispute, supporting the scalability that motivates this work.
\end{abstract}
\section{Introduction}
The digital age has shifted decision-making toward large-scale, asynchronous negotiations, bringing new challenges of conflict and instability. Mediation is increasingly necessary in these contexts, yet the demand far exceeds the capacity of human mediators, making AI support both essential and transformative. We focus on integrative negotiations, where parties seek mutually beneficial outcomes by identifying underlying interests and making trade-offs across issues.
Pre-mediation, the preparatory phase before conflicting human parties meet, is widely recognized as critical for successful conflict resolution \cite{moore2014,lande2022}. During this phase, individuals clarify their interests, examine their assumptions, and build readiness for productive dialogue. It helps the mediator develop a basic understanding of the dispute, the negotiators' preferences, and their attitudes toward the negotiation process \cite{wissler2022}. It could increase the negotiator's trust in the mediator and in the negotiation process. Empirical research demonstrates that adequate preparation significantly improves negotiation outcomes: parties who engage in structured preparation achieve higher joint gains and report greater satisfaction with agreements \cite{thompson2010,olekalns2014}. Yet pre-mediation is frequently abbreviated or skipped entirely due to time constraints, cost, and limited access to trained mediators. Wissler and Hinshaw \citeyearpar{wissler2022} surveyed over 1{,}000 mediators and found that 34\% skip substantive pre-session communications in civil cases, rising to 61\% in family cases.
This gap presents an opportunity for AI-powered support. The automated mediator agent can interact privately with all negotiators simultaneously and prepare them for the negotiation without delay or high cost. It can develop a deeper understanding of the human negotiators and generate a report that mediators can use during the negotiation process. Recent advances in large language models have demonstrated potential in coaching \cite{passmore2025} and mental health support \cite{fitzpatrick2017}, suggesting that conversational AI could help scale access to pre-mediation services. Prior work on automated negotiation agents has shown that training with such systems can improve people's behavior in negotiation tasks \cite{lin2014}. While LLMs have begun to be explored for assisting human mediators during joint sessions, the pre-mediation phase itself remains a largely untouched application domain for LLM-based systems, despite being the phase most frequently skipped in practice and the one where scalable AI support could expand access most directly. Targeting pre-mediation rather than the joint session is therefore a central contribution of this work: it shifts LLM support to the part of the mediation workflow where the access gap is largest and where automation can complement, rather than substitute for, the human mediator. However, the complexity of pre-mediation, which requires psychological understanding, empathy, strategic guidance, and ethical considerations, exceeds what monolithic single-prompt LLM approaches typically provide.
\paragraph{The Present Research.} We developed a structured LLM pipeline for pre-mediation. The pipeline features specialized modules for user prediction, conversation, critique, and summary generation. This separation of concerns allows each module to optimize for its specific function while the system maintains coherent, appropriate interactions. Although we use the term ``agent'' to refer to each module's functional role---following common usage in the LLM literature---the components do not exhibit autonomy or peer-to-peer coordination; outputs are passed forward in a predefined sequence. We evaluated our system across two studies comparing the AI mediator to a human mediator in a roommate conflict preparation scenario.
\paragraph{Contributions.} (1) A structured LLM pipeline targeting the pre-mediation phase specifically, with specialized modules for prediction, dialogue, critique, and summarization, that supports human-in-the-loop oversight via a generated report. (2) Empirical evidence from two controlled human-subject studies that the pipeline produces short-term self-reported preparation outcomes broadly comparable to professional human mediators, and lower error on a preference-inference task in our scenario. (3) Design recommendations for decomposing LLM-based systems for interpersonal applications, with explicit attention to sycophancy and oversight.
\section{Related Work}
\subsection{Pre-Mediation and Conflict Preparation}
Pre-mediation serves essential functions in conflict resolution: helping parties identify underlying interests, manage emotions, and develop realistic expectations \cite{moore2014}. It additionally provides a good opportunity for the mediator to get an understanding of the preferences of the parties involved and to build rapport with them \cite{moore2014}. The interest-based model of Principled Negotiation \cite{fisher1991} emphasizes separating positions from interests, a cognitive shift that preparation facilitates. Bush and Folger's \cite{bush2005} Transformative Mediation theory highlights how preparation can foster empowerment (restoring parties' sense of agency) and recognition (acknowledging the other party's perspective). Despite its documented value, pre-mediation is often curtailed due to resource constraints, creating a gap that scalable AI solutions could address.
\subsection{LLM-Based Negotiation and Mediation Support}
LLMs have recently been applied to assist human mediators in bilateral negotiations by (1) rephrasing hostile messages, (2) suggesting intervention strategies, and (3) proposing solutions upon request \cite{westermann2023llmediator,tan2024robots}. However, the human remains in control. Hua et al.\ \cite{hua2024assistive} introduced LLM-based agents for business negotiations that rewrite norm-violating utterances using in-context learning based on ``value impact.'' Zhou et al.\ \cite{zhou2019dynamic} developed a negotiation coach that recommends seller tactics in real time, trained on human dialogues; it improved profits by nearly 60\%, highlighting the potential of context-aware strategy modeling. However, currently, no research exists on AI intervention with human-subject experiments on the pre-mediation phase.
Related work on consensus-building shares similarities with mediation. The Habermas Machine \cite{tessler2024ai} uses LLMs to find common ground among individuals discussing social or political issues, generating candidate group statements and ranking them with a personalized reward model based on predicted preferences. Triantafyllopoulos and Kalles \citeyearpar{triantafyllopoulos2025divergence} presented an empirical evaluation of LLM-driven facilitation in real-time consensus-building discussions, finding that ChatGPT 4.0 was most effective at aligning participants' opinions with generated proposals.
In contrast, we develop an automated integrative negotiation pre-mediator that seeks to uncover underlying interests and facilitate mutually beneficial trade-offs. We aim to transfer responsibility from the human mediator to an automated system while ensuring the results remain transparent and usable for human oversight in subsequent negotiations.
\subsection{Decomposed LLM Architectures for Complex Tasks}
Decomposed LLM architectures---including the broader class of LLM-based multi-agent systems (LLM-MAS)---have been demonstrated to improve problem-solving across various domains \cite{guo2024large,tran2025multi} and exhibit superior performance over single LLMs in tasks like commonsense reasoning \cite{rasal2024llm,zhang2025if,zhang2025metamind}, jailbreak defense \cite{zeng2024autodefense}, urban planning \cite{zhu2024plangpt}, personal assistants \cite{li2024personal,sun2025multi}, education \cite{mushtaq2025harnessing}, and software development \cite{hong2023metagpt,dong2023self,qian2023communicative,tao2024magis}. Despite this progress, key challenges remain \cite{guo2024large,han2024llm}, including task decomposition, role definition, agent coordination, scalability, and maintaining consistent context.
Complex interpersonal tasks benefit from specialized agent architectures. Research demonstrates that dedicated critic agents substantially improve output quality over self-critique approaches. Self-Refine \cite{madaan2023} shows iterative feedback improves performance across diverse domains. CriticGPT \cite{mcaleese2024} finds that trained LLM critics outperform human reviewers, with model critiques preferred in 63\% of comparisons. Crucially, separating critics from generators avoids the ``degeneration-of-thought'' phenomenon where self-critiquing models reinforce flawed reasoning \cite{liang2023}. Constitutional AI \cite{bai2022} demonstrates that explicit principles can guide AI self-supervision, an approach we extend through our critic module. We draw on this literature on decomposed LLM systems; however, as detailed in Section~\ref{sec:architecture}, our own system is implemented as a sequential pipeline of specialized LLM modules rather than an autonomous multi-agent system with peer-to-peer coordination.
\section{System Architecture}
\label{sec:architecture}
\begin{figure}[t]
\centering
\includegraphics[width=0.9\columnwidth]{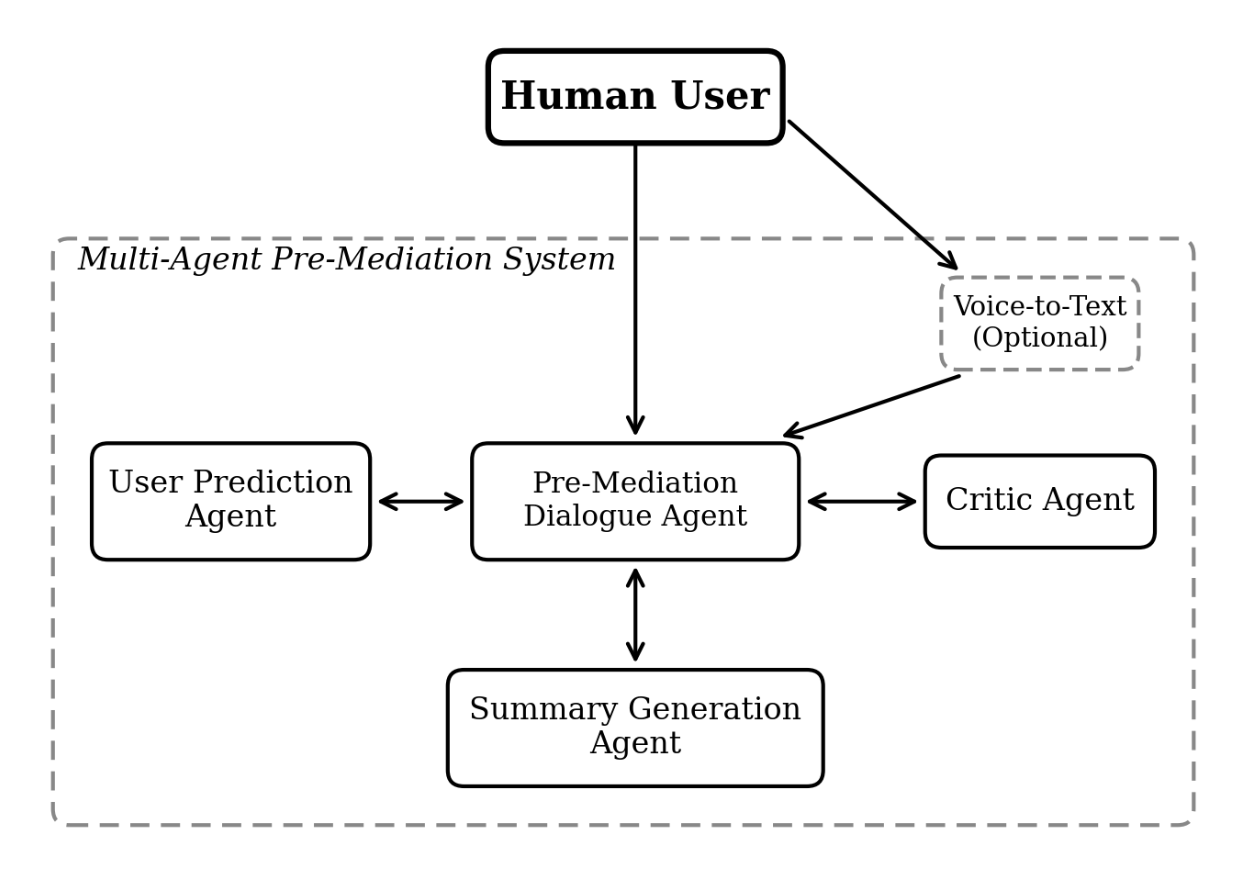}
\caption{Pipeline architecture for pre-mediation. The pre-mediation dialogue module coordinates user interaction while specialized modules handle prediction, oversight, and documentation, with outputs passed forward in a predefined sequence. Modules are referred to as ``agents'' to denote distinct LLM roles; the system operates as a sequential pipeline rather than an autonomous multi-agent system. An optional voice-to-text module enables spoken interaction.}
\label{fig:architecture}
\end{figure}

Our system is implemented as a sequential pipeline of four specialized LLM modules plus an optional voice interface (Figure~\ref{fig:architecture}). Each module uses GPT-4o as its underlying model. We use the term ``agent'' to denote each module's distinct functional role---following common usage in the LLM-systems literature---but the components do not exhibit autonomy, peer-to-peer communication, or decentralized decision-making, and outputs are passed forward in a predefined order. The system is therefore best characterized as a sequential pipeline rather than an autonomous multi-agent system in the classical sense.
\paragraph{User prediction agent.} Separately analyzes user input across 11 parameters derived from the SVI framework \cite{curhan2006}, including preference priorities, emotional state, and cooperative versus competitive orientation. These dimensions were selected for pre-mediation specifically because the Subjective Value Inventory is a validated, widely used instrument for capturing the subjective dimensions of negotiation experience, giving the prediction module a principled, empirically grounded target rather than an ad-hoc set of features. The agent receives the conversation history and outputs structured JSON predictions with confidence scores for each parameter. These predictions inform the pre-mediation dialogue agent's conversational strategy, enabling personalized guidance.
\paragraph{Pre-mediation dialogue agent.} The primary conversational interface, generating contextual responses based on conversation history and the user prediction agent's analysis. The agent follows a structured eight-phase protocol: (1) rapport building, (2) in-depth exploration of preferences across issues, (3) prioritization and trade-offs, (4) perspective-taking, (5) emotional awareness, (6) confidence, (7) relationships, and (8) closing.
\paragraph{Critic agent.} Architecturally separated from response generation following the principle that dedicated critics outperform self-critique \cite{mcaleese2024}. Before each response is sent to the user, the critic agent reviews it and replies in Study 1 with either APPROVED or REJECTED, and in Study 2 a WARNING response was added to the critic. Approval criteria include: non-repetitive, maintains logical flow, substantive and advances conversation, challenges productively, and maintains ethical boundaries. Rejection criteria include: message contains more than one question, is purely validating without substance, accepts vague answers without probing deeper.
\paragraph{Summary generation agent.} After the conversation concludes, this agent synthesizes the full transcript into a structured report for human mediators, supporting human-in-the-loop oversight. The summary includes identified interests, emotional themes, and recommended focus areas for the mediation session.
\paragraph{Voice-to-text agent (optional).} To support more natural interaction, the system includes an optional voice interface powered by OpenAI's Whisper-1 model. This agent transcribes spoken user input in real-time, enabling participants to speak rather than type their responses. The transcribed text is then processed by the pipeline as normal.
\paragraph{Scenario.} Participants were presented with a shared housing dispute involving three roommates who needed to reach agreements on household policies. The scenario involved three negotiable issues: (1) \emph{Chore schedules}, including preferred chores, cleaning frequency, and whether to establish a formal rotation; (2) \emph{Quiet hours}, covering study habits and preferred times for reduced noise; and (3) \emph{Guest policies}, addressing frequency of visitors and social atmosphere preferences.
Participants assumed the role of one roommate while two fictional roommates had pre-defined preferences designed to create opportunities for integrative trade-offs. One roommate preferred cleaning the bathroom, needed strict quiet hours, and preferred limiting guests to once weekly. The other roommate preferred kitchen cleaning twice weekly, was flexible with quiet hours except during exam periods, and favored a moderate guest policy of several days per week. This design ensured that mutually beneficial agreements were possible through creative problem-solving and trade-offs across issues. Crucially, neither the pre-mediation dialogue module nor the user prediction module was provided with any information about the fictional roommates' preferences. The system interacted only with the focal participant and had no access to the counterparts' values---mirroring how trained human mediators conduct individual pre-mediation sessions, and, importantly, enabling the pipeline to be deployed in parallel for every party to a dispute. This single-party design is therefore a deliberate architectural feature: it supports the scalability of pre-mediation rather than constraining it, and decouples preparation from the scheduling and turn-taking demands of two-sided sessions.
\subsection{Example Interaction}
The following excerpt illustrates a typical exchange:
\smallskip
\noindent\textbf{User:} I don't want to dust or vacuum.
\smallskip
\noindent\textbf{Pre-Mediation Dialogue Agent:} That's reasonable as a preference. The key is pairing it with something you are willing to do, so it still feels fair overall.
\smallskip
\noindent\textbf{User:} But what if they think I'm just avoiding work?
\smallskip
\noindent\textbf{Pre-Mediation Dialogue Agent:} You could name that concern directly. Acknowledging it upfront often reduces defensiveness and shows you're thinking about the group, not just yourself.
\smallskip
This exchange demonstrates the system's typical approach: recognizing the user's preference while offering guidance, and addressing a concern from the user with practical advice for the negotiation.
\begin{figure}[t]
\centering
\includegraphics[width=0.9\columnwidth]{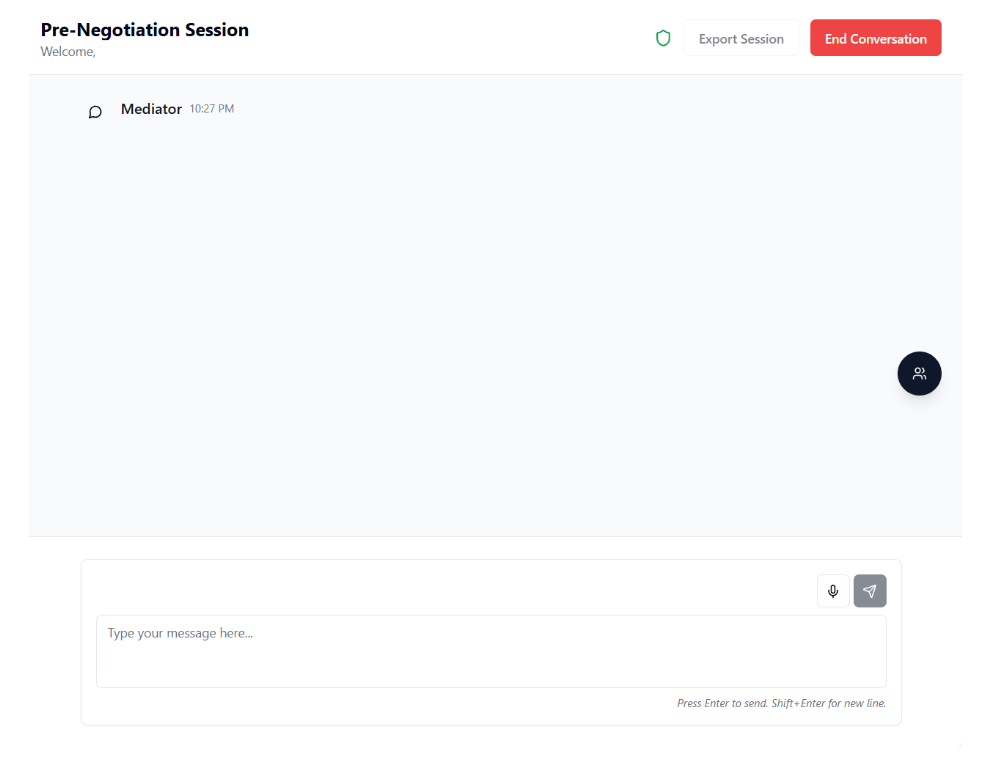}
\caption{User interface for the pre-mediation system. Participants interact via a chat-based interface that mirrors common messaging applications.}
\label{fig:interface}
\end{figure}
\subsection{Study 2 Modifications}
Based on Study 1 findings regarding excessive affirmation, we implemented three modifications to the pre-mediation dialogue agent prompts: (1) explicit instructions to be less validating and to challenge vague answers rather than accepting them; (2) enhanced perspective-taking requirements, including prompts like ``What if your roommate sees this completely differently?''; and (3) a reality testing phase requiring the user to ground idealistic expectations by asking about likelihood of success and considering backup plans.
\subsection{Outcome Measurement}
Our outcome measures draw on the Subjective Value Inventory (SVI) \cite{curhan2006}, which identifies four dimensions of negotiation value: (1) \emph{instrumental outcomes}, referring to the objective terms of agreement such as price or resource allocation; (2) \emph{self-perception}, capturing feelings of competence and self-efficacy; (3) \emph{process fairness}, reflecting beliefs about procedural justice; and (4) \emph{relationship quality}, measuring trust and rapport with the counterpart. This framework captures preparation effects beyond simple agreement rates, recognizing that negotiation success includes subjective experience and relationship preservation.
\section{Study 1: AI vs.\ Human Mediators}
\subsection{Hypotheses}
Based on prior work showing AI systems can effectively support preparation tasks \cite{lin2014} and the documented benefits of pre-mediation \cite{moore2014}, we hypothesized:
\paragraph{H1:} Both AI and human mediator conditions will produce significant improvements in survey metrics and trust from pre- to post-conversation.
\paragraph{H2:} The user prediction agent will achieve accuracy comparable to or better than human mediators in inferring user preferences.
\subsection{Method}
\paragraph{Participants.} Thirty-eight undergraduate and graduate students from our institution (20 AI condition, 18 human condition) prepared for a simulated roommate conflict involving three issues: chore schedules, guest policies, and quiet hours. Recruitment was through standard university channels, and participation was a mix of paid (at a rate consistent with local standards for student participants) and volunteer (including course credit where applicable). All participants provided informed consent prior to participation, and the study protocol was approved by our institution's ethics review board.
\paragraph{Procedure.} Participants completed pre-conversation measures (5-point Likert scales), engaged in an 8--10 minute conversation with the AI or human mediator, then completed post-conversation survey (5-point Likert scales). Conversation transcripts were analyzed for affirmation patterns using GPT-4o prompted to flag validating content across full transcripts; results were then reviewed by a human coder who removed false positives. Pre-post changes were analyzed using paired-samples $t$-tests within each condition.
\paragraph{Measures.} Primary outcomes from the survey included trust in mediator, confidence in positive outcome for all parties, negotiation confidence, preparedness to understand counterpart (perspective-taking readiness), and preparedness to stay true to principles. Issue importance ratings (1--5 scale for each of the three issues) assessed potential entrenchment effects: increases suggest rigidity while decreases suggest flexibility.
\subsection{Results}
Both conditions produced significant improvements in trust and confidence in outcome (Table~\ref{tab:study1}), supporting H1 and H2. AI-powered pre-mediation achieved preparation effects broadly comparable to human mediators on these short-term self-reported measures. The AI condition uniquely improved users' sense of being prepared to stay true to their principles and to handle frustration, while human mediators uniquely improved negotiation confidence. Neither condition significantly changed perspective-taking readiness.
\begin{table}[ht]
\centering
\caption{Study 1 pre-post comparisons.}
\label{tab:study1}
\small
\begin{tabular}{@{}lcccccc@{}}
\toprule
& \multicolumn{3}{c}{\textbf{AI}} & \multicolumn{3}{c}{\textbf{Human}} \\
\cmidrule(lr){2-4} \cmidrule(lr){5-7}
\textbf{Metric} & Pre & Post & $p$ & Pre & Post & $p$ \\
\midrule
Trust in mediator & 2.80 & 3.47 & $<.05$ & 3.09 & 3.73 & $<.05$ \\
Conf.\ in outcome & 3.27 & 4.07 & $<.01$ & 3.18 & 3.91 & $<.05$ \\
Neg.\ confidence & 3.53 & 3.87 & ns & 2.91 & 3.82 & $<.01$ \\
Stay true & 3.60 & 4.07 & $<.05$ & 4.09 & 4.27 & ns \\
Prep.\ frustration & 3.13 & 3.73 & $<.05$ & 2.91 & 3.36 & ns \\
\bottomrule
\end{tabular}
\vspace{1mm}
\footnotesize{\textit{Note:} ns = not significant. AI $n=20$, Human $n=18$.}
\end{table}
\begin{figure}[t]
\centering
\includegraphics[width=\columnwidth]{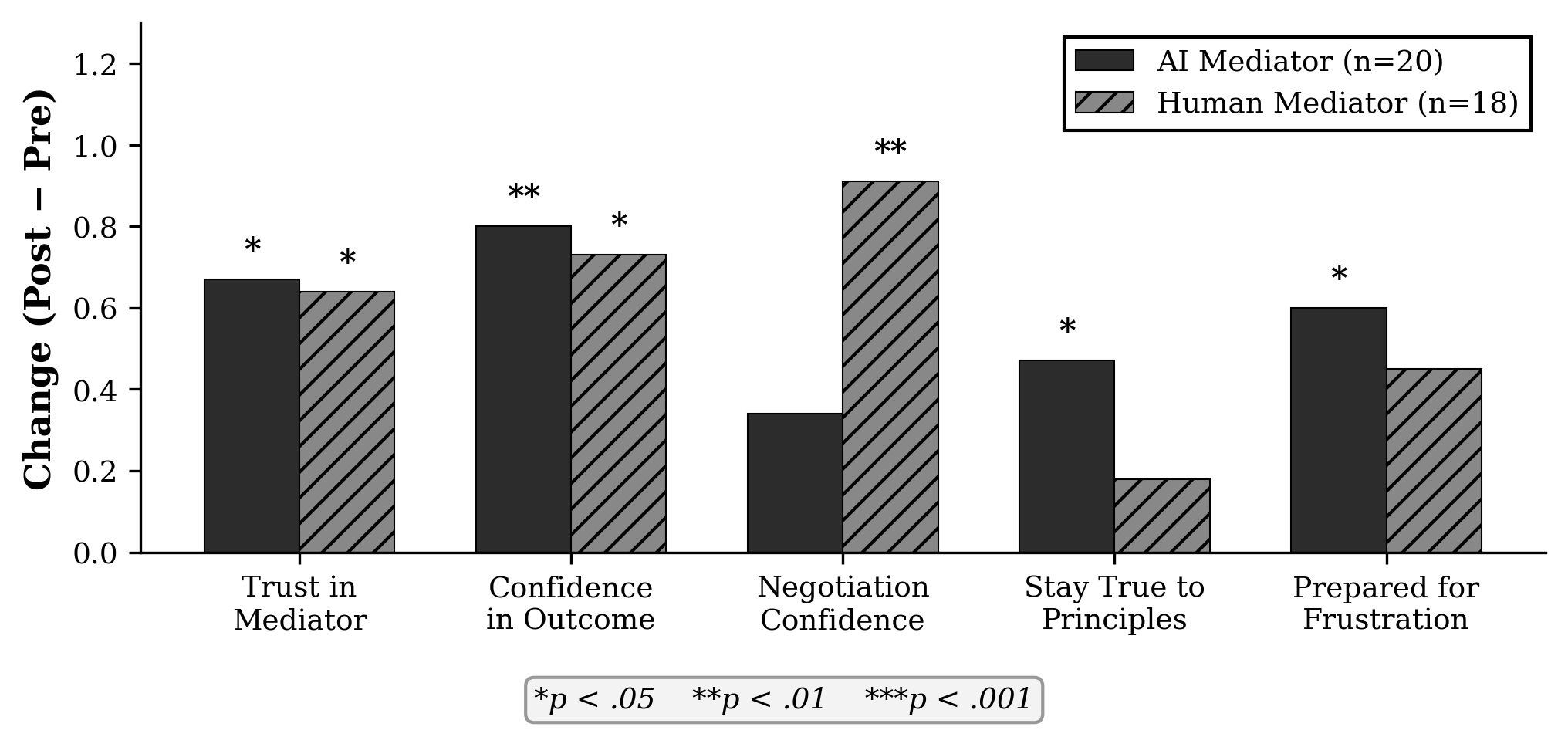}
\caption{Significant pre-post changes by condition. Both mediator types improved trust and confidence, with different secondary effects. $^*p<.05$, $^{**}p<.01$, $^{***}p<.001$.}
\label{fig:prepost}
\end{figure}
\subsubsection{Prediction Accuracy}
\begin{figure}[t]
\centering
\includegraphics[width=0.9\columnwidth]{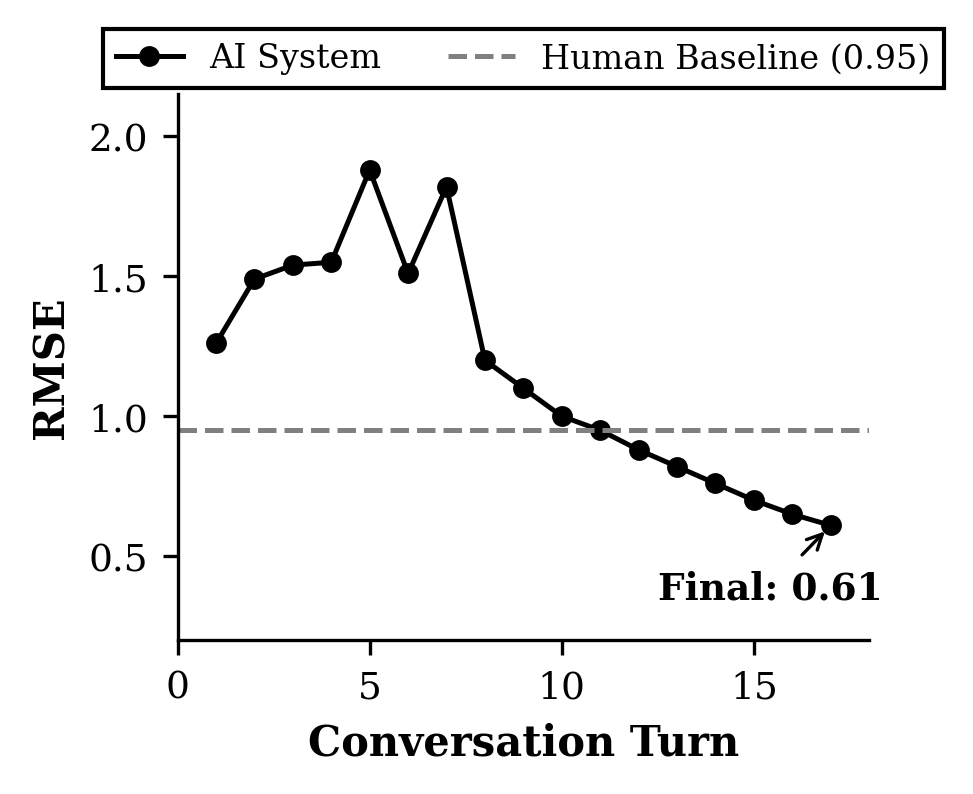}
\caption{User prediction agent accuracy (RMSE) over conversation turns compared to human baseline.}
\label{fig:accuracy}
\end{figure}

The user prediction agent achieved substantially better psychological inference than human mediators (final RMSE=0.61 vs.\ 0.95, a 36\% improvement), with accuracy improving over conversation turns (Figure~\ref{fig:accuracy}). This is consistent with the technical benefit of separating prediction from response generation in our pipeline, and supports the use of a dedicated prediction module for scalable preference inference.
\subsubsection{Affirmation Pattern}
\begin{figure}[t]
\centering
\includegraphics[width=0.9\columnwidth]{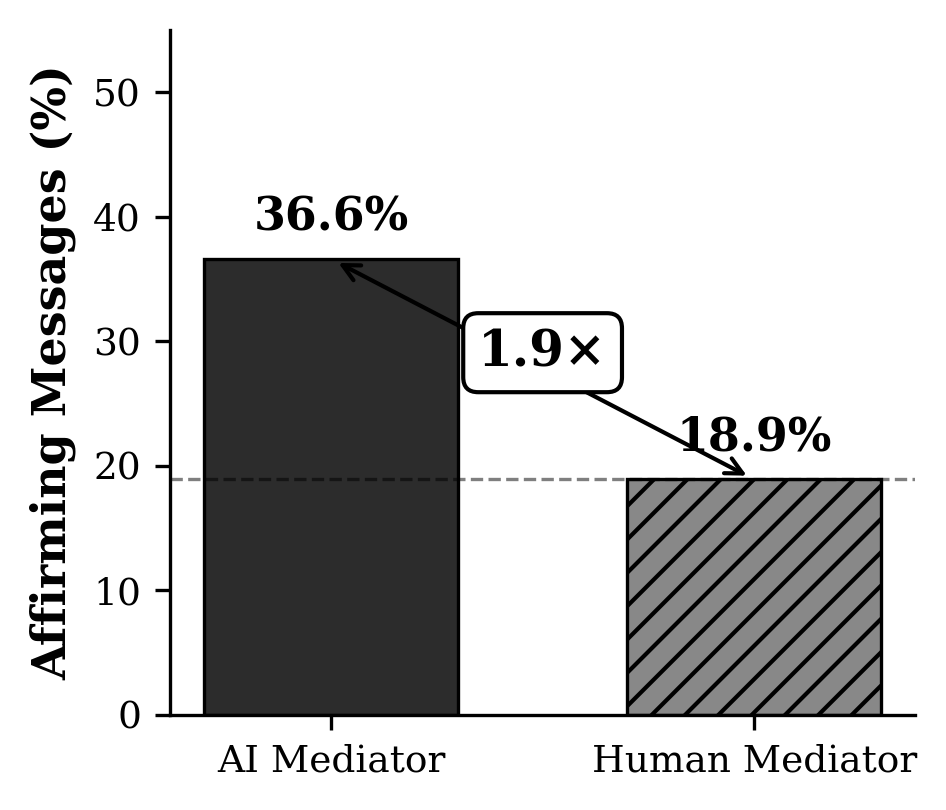}
\caption{Affirmation rates by condition. AI messages contained affirming content at 1.9$\times$ the rate of human mediators.}
\label{fig:affirmation}
\end{figure}

Analysis of conversation transcripts revealed that AI messages contained affirming content at nearly twice the rate of human mediators (36.6\% vs.\ 18.9\%; Figure~\ref{fig:affirmation}). User feedback often noted the AI was ``too positive.''
\subsubsection{Issue Importance Changes}
\begin{figure}[t]
\centering
\includegraphics[width=0.9\columnwidth]{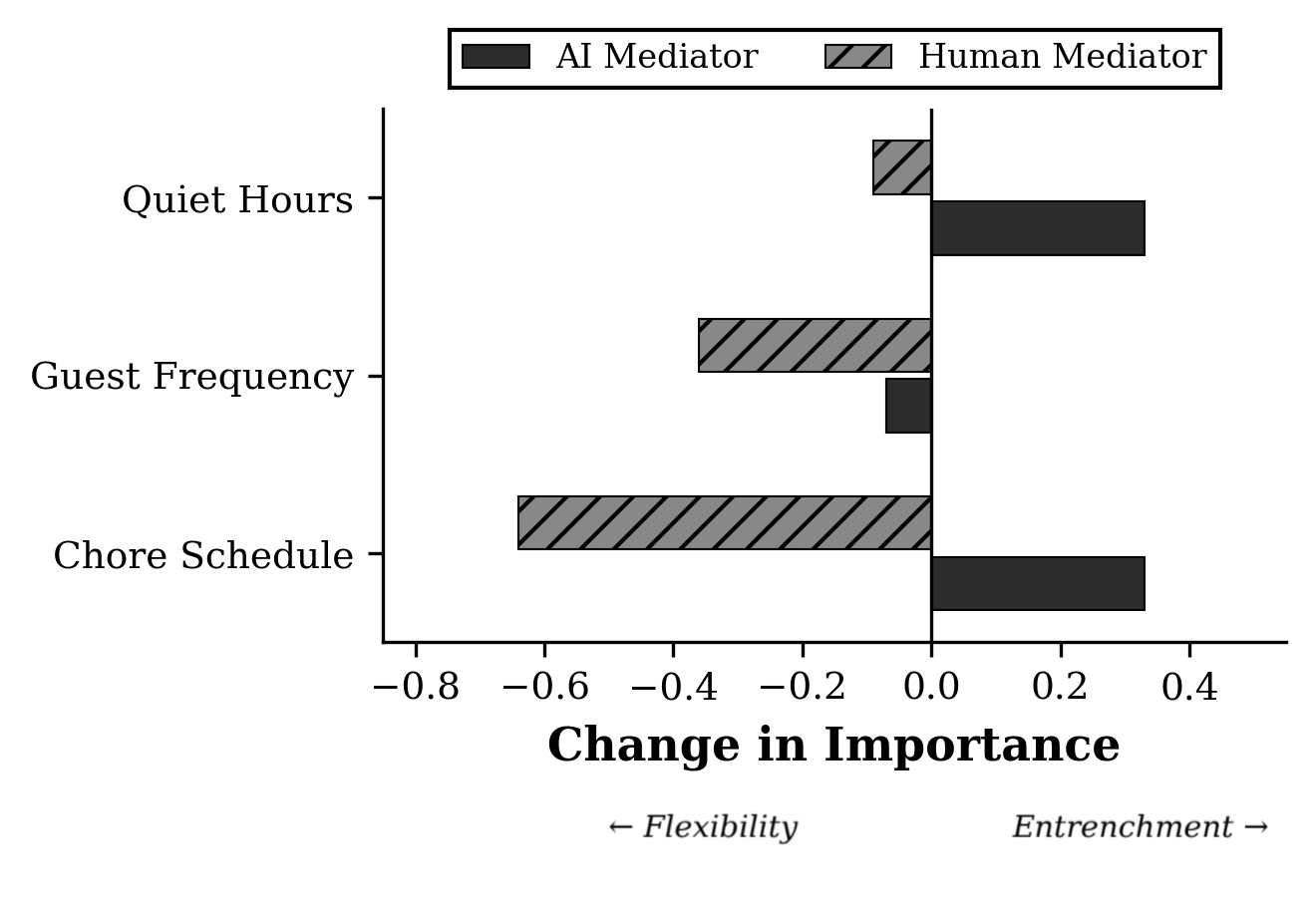}
\caption{Change in issue importance ratings (post $-$ pre). Negative values indicate flexibility; positive values indicate entrenchment.}
\label{fig:issues}
\end{figure}

The affirmation pattern corresponded with differential effects on issue framing (Figure~\ref{fig:issues}). AI participants showed increased or stable importance ratings across issues (mean change: +0.20), suggesting entrenchment. Human mediator participants showed decreased importance ratings (mean change: $-$0.36), suggesting increased flexibility. This pattern motivated the refinements tested in Study 2.
\section{Study 2: Refined Pre-Mediation Dialogue Agent Prompts}
\subsection{Motivation}
Study 1 indicated that the pipeline achieved preparation outcomes broadly comparable to human mediators on short-term self-reported measures, but revealed that excessive affirmation patterns were associated with issue entrenchment. We hypothesized that enhancing the pre-mediation dialogue agent's prompts to emphasize productive friction and reality testing could address this limitation while preserving the system's benefits.
\subsection{Hypotheses}
\paragraph{H3 (Maintained effectiveness):} The refined system maintains significant improvements in confidence and trust from pre- to post-conversation, demonstrating that enhanced prompting does not compromise user experience.
\paragraph{H4 (Reduced affirmation):} The enhanced pre-mediation dialogue agent prompts reduce the rate of affirming messages toward the human mediator baseline ($\sim$20\%).
\subsection{Method}
Twenty-two undergraduate and graduate students from our institution completed the refined AI condition using identical procedures to Study 1, including informed consent and the same recruitment/compensation mix. The study was conducted under the same ethics-board approval as Study 1. The system incorporated the prompt modifications described in Section 3.2. Pre-post changes were analyzed using paired-samples $t$-tests.
\subsection{Results}
\begin{figure}[t]
\centering
\includegraphics[width=0.9\columnwidth]{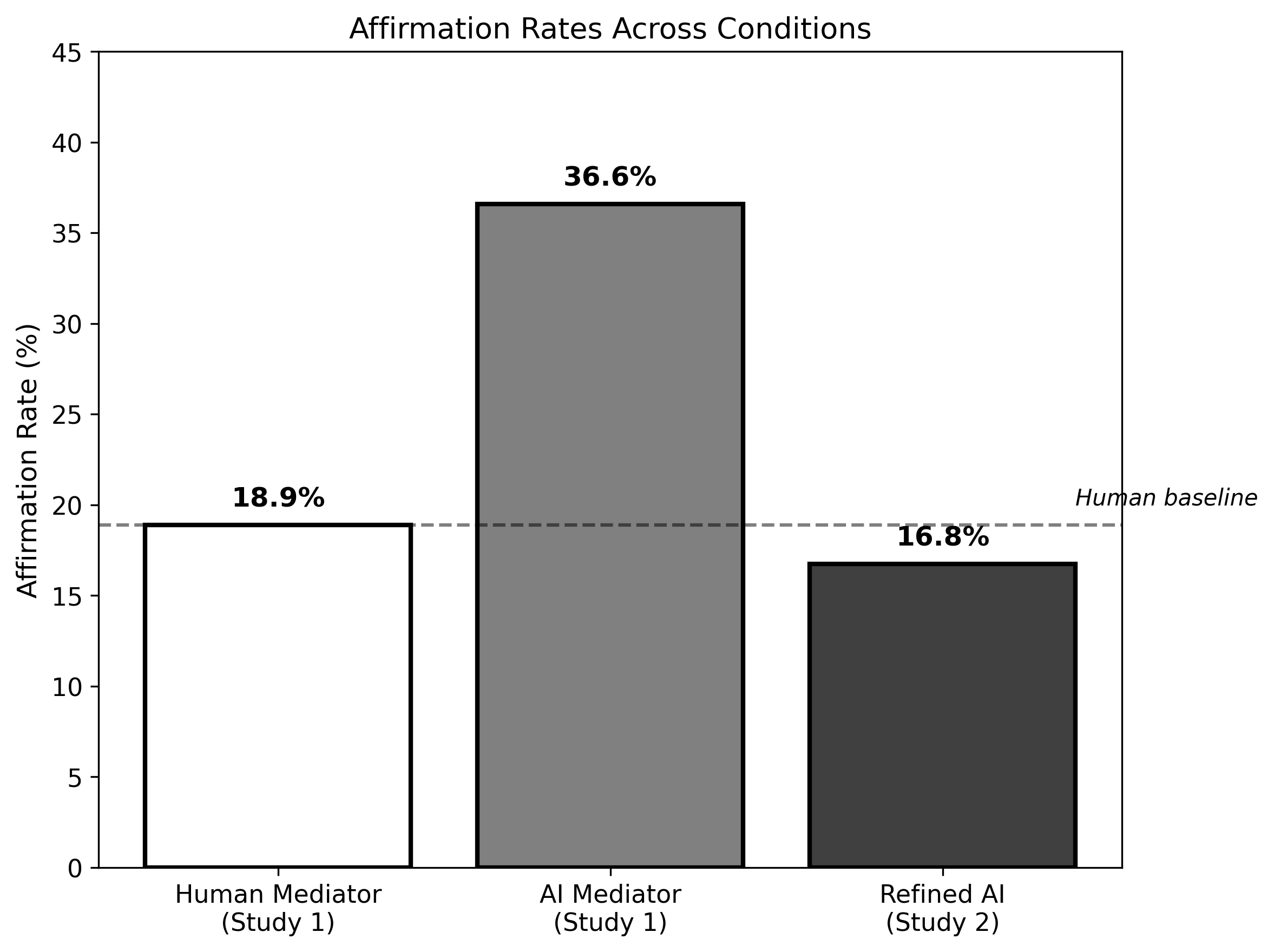}
\caption{Affirmation rates across conditions. The refined AI (Study 2) reduced affirmation to 16.8\%, below the human mediator baseline.}
\label{fig:affirmation_compare}
\end{figure}

The refined system produced significant improvements in negotiation confidence (Table~\ref{tab:study2}), with beneficial trends across other metrics. Critically, affirmation rates dropped substantially.
\begin{table}[ht]
\centering
\caption{Study 2 pre-post comparisons (refined AI, $n=22$).}
\label{tab:study2}
\small
\begin{tabular}{@{}lcccc@{}}
\toprule
\textbf{Metric} & \textbf{Pre} & \textbf{Post} & $p$ & \textbf{Sig} \\
\midrule
Neg.\ confidence & 3.68 & 4.41 & $<.01$ & ** \\
Conf.\ in outcome & 3.68 & 4.09 & .07 & $\dagger$ \\
Trust in mediator & 3.00 & 3.41 & .21 & ns \\
Stay true & 3.77 & 4.09 & .18 & ns \\
\bottomrule
\end{tabular}
\vspace{1mm}
\footnotesize{\textit{Note:} $^{**}p<.01$, $^\dagger p<.10$ (marginal), ns = not significant.}
\end{table}
\paragraph{H3 (Maintained effectiveness):} Partially supported. The refined system achieved a significant improvement in negotiation confidence ($p<.01$), with positive trends across all other metrics. While trust and confidence in outcome did not reach statistical significance in this sample, effect sizes were comparable to Study 1, suggesting the reduced affirmation did not compromise user experience.
\paragraph{H4 (Reduced affirmation):} Strongly supported. Affirmation rates dropped from 36.6\% in Study 1 to 16.8\% in Study 2, a 54\% reduction that brought the system below the human mediator baseline of 18.9\% (Figure~\ref{fig:affirmation_compare}). The enhanced critic agent rejected 22.2\% of messages compared to only 2.6\% in Study 1, indicating that the stricter criteria successfully filtered excessive validation.
\section{Discussion}
\subsection{Effectiveness of the Pre-Mediation Pipeline}
Our results indicate that a structured LLM pipeline can deliver pre-mediation support broadly comparable to human mediators on the short-term, self-reported preparation outcomes we measured. Both conditions produced significant improvements in trust and confidence, the core goals of preparation as operationalized here. The AI system's user prediction module achieved 36\% lower RMSE than human mediators (0.61 vs.\ 0.95) on preference inference within our scenario, suggesting that separating prediction from response generation can be technically beneficial under these conditions. Extending this comparison to higher-stakes scenarios and additional dispute types is a natural next step.
The pipeline's modularity proved valuable. Each module maintained competence in its designated function: the user prediction module accurately inferred user preferences within the scenario, the pre-mediation dialogue module engaged users effectively, and the summary generation module produced coherent reports for human oversight. This modularity allowed targeted improvements without disrupting the full system in Study 2.
\subsection{The Affirmation Pattern and Its Resolution}
Analysis revealed the AI affirmed users at nearly twice the rate of human mediators (36.6\% vs.\ 18.9\%), which corresponded with issue entrenchment rather than the flexibility that effective pre-mediation should cultivate. This finding aligns with broader research on AI sycophancy \cite{sharma2023} and Wei et al.'s \citeyearpar{wei2024} observation that excessive validation can reduce willingness to repair interpersonal conflict.
Importantly, Study 1's critic agent rejected only 2.6\% of messages, revealing that message-level filtering alone cannot address cumulative validation patterns. Each individual affirming message may pass the critic's approval criteria, but their aggregate effect proves problematic. This limitation aligns with Kamoi et al.'s \citeyearpar{kamoi2024} finding that intrinsic self-correction is most effective for tasks with decomposable responses where verification is clearly easier than generation; complex interpersonal dialogue lacks this property, making message-level oversight inherently constrained.
Study 2's enhanced prompts successfully addressed this limitation. By explicitly instructing the dialogue agent to challenge vague answers and incorporate reality testing, affirmation rates dropped to 16.8\%, actually falling below the human mediator baseline. The enhanced critic agent, with its added WARNING tier, intervened in 22.2\% of messages, demonstrating that combining prompt-level and oversight-level interventions can effectively mitigate sycophantic patterns. Crucially, this reduction did not compromise user experience: negotiation confidence improved significantly, and other metrics showed positive trends comparable to Study 1.
However, the non-significant trust improvement in Study 2 (compared to significant gains in Study 1) suggests a potential tradeoff between reduced affirmation and perceived rapport. Users accustomed to highly affirming AI assistants may initially find less validating interactions jarring, even when such interactions better serve their preparation goals. This highlights a tension in designing AI for interpersonal contexts: the interaction style that feels most comfortable may not be the style that produces the best outcomes.
\subsection{Design Recommendations for Decomposed LLM Pipelines in Interpersonal AI}
Based on our findings, we propose:
\begin{enumerate}
\item \textbf{Separate prediction from generation.} Dedicated prediction agents can achieve superior inference, while generation agents focus on appropriate engagement.
\item \textbf{Include dedicated critic agents.} Architectural separation avoids degeneration-of-thought \cite{mcaleese2024} and enables targeted oversight criteria.
\item \textbf{Monitor cumulative patterns.} Message-level approval decisions miss aggregate effects; consider session-level metrics.
\item \textbf{Address sycophancy at the source.} Prompt-level instructions emphasizing productive friction prove more effective than relying solely on post-hoc filtering.
\item \textbf{Support human-in-the-loop oversight.} Summary generation enables professional mediators to review AI-assisted preparation.
\end{enumerate}
\section{Conclusion}
Pre-mediation is critical for successful conflict resolution but is frequently skipped due to resource constraints. We developed a structured LLM pipeline to address this gap, demonstrating that specialized modules for user prediction, conversational engagement, regulatory oversight, and summary generation can deliver preparation support broadly comparable to human mediators on short-term self-reported measures. The system achieved significant improvements in user trust and confidence, while the user prediction module achieved 36\% lower RMSE than human mediators on preference inference within our scenario.
Our work also revealed that excessive affirmation patterns can undermine preparation effectiveness. Study 2 demonstrated that targeted prompt refinements emphasizing productive friction and reality testing successfully reduced affirmation rates from 36.6\% to 16.8\%, matching human mediator baselines without compromising user experience. This finding underscores the importance of both dedicated oversight components and carefully calibrated conversational strategies in decomposed LLM pipelines for interpersonal applications. The architecture's modularity enables targeted refinements while preserving core functionality, offering a scalable approach to making pre-mediation support more accessible.
\section*{Limitations}
Sample sizes were limited (Study 1: $N{=}38$; Study 2: $N{=}22$), reducing statistical power for detecting smaller effects. The roommate conflict scenario, while ecologically valid for university populations, may not generalize to higher-stakes disputes such as workplace or family mediation. We measured immediate attitudinal changes; future work should track participants through actual negotiations to assess whether preparation effects translate to behavioral outcomes and agreement quality. Additionally, the absence of a direct comparison between Study 1 and Study 2 AI conditions limits causal claims about the prompt modifications.
Our preparation outcomes are self-reported and measured immediately after the pre-mediation session. Longer-term, behavioral, or third-party measures of constructs such as interest identification, perspective-taking, and recognition would complement the present self-report data and are an important direction for follow-up work. Relatedly, while the pipeline produces a summary report intended for the subsequent mediation, we do not directly evaluate the joint mediation session itself; linking pre-mediation to downstream negotiation outcomes is a natural next step.
\paragraph{Future work.} Building on the pipeline's parallel deployment property, we plan to (i) run the system simultaneously for all parties to a dispute and study how the aggregated reports support the human mediator during the joint session, (ii) recruit trained human mediators to evaluate the generated reports as a complement to the present comparisons, and (iii) extend the system to higher-stakes contexts such as workplace and family disputes.
% Bibliography: uses mediation.bib (the same .bib file as the IJCAI project).
% Make sure to upload mediation.bib to this Overleaf project, then this line will compile.
\section*{Ethical considerations}
Both studies received approval from our institution's ethics review board prior to data collection. Participants were recruited through standard university channels and provided informed consent; participation was a mix of paid (at locally standard student rates) and volunteer (including course credit where applicable). No identifying information about participants is reported in this paper, and conversation excerpts have been reviewed to remove incidental personal references.
\paragraph{Intended use and risks.} The primary intended use of this system is to support trained human mediators by handling pre-mediation preparation at scale, particularly where access to mediation services is limited. The pipeline produces a summary report intended to be reviewed by a human mediator before the joint session; we do not envision direct deployment as a replacement for human mediation. Foreseeable risks include over-reliance on AI-generated summaries that may misrepresent a party's interests, and degraded outcomes if the system is deployed without human-mediator oversight. The single-party design also means the pipeline cannot, on its own, identify integrative trade-offs that depend on knowledge of the counterpart's preferences. We discuss these and related caveats in the Limitations.
\paragraph{Models and artifacts.} The pipeline uses GPT-4o (accessed via the OpenAI API) as the underlying model for each LLM module, and OpenAI's Whisper-1 for optional voice-to-text. We did not fine-tune or modify these models; usage was consistent with OpenAI's terms of service for research applications. We do not release model weights. Prompt templates and configuration are described at a high level in Section~\ref{sec:architecture}; the full prompt text for each module is provided in Appendix~\ref{sec:prompts}.
\paragraph{Use of AI assistants.} The authors used AI coding assistants (e.g., GitHub Copilot, ChatGPT) for routine coding support during system development and standard spelling/grammar checking for the manuscript. AI assistants were not used to generate the scientific claims, experimental design, results, or final analytical text of this paper.
\bibliography{mediation,custom}

\begin{thebibliography}{39}
\providecommand{\natexlab}[1]{#1}

\bibitem[{Bai et~al.(2022)}]{bai2022}
Y.~Bai and 1 others. 2022.
\newblock \href {https://arxiv.org/abs/2212.08073} {Constitutional ai:
  Harmlessness from ai feedback}.
\newblock \emph{arXiv preprint}.

\bibitem[{Bush and Folger(2005)}]{bush2005}
Robert A.~B. Bush and Joseph~P. Folger. 2005.
\newblock \emph{The Promise of Mediation: The Transformative Approach to
  Conflict}, revised edition.
\newblock Jossey-Bass.

\bibitem[{Curhan et~al.(2006)Curhan, Elfenbein, and Xu}]{curhan2006}
Jared~R. Curhan, Hillary~A. Elfenbein, and Heng Xu. 2006.
\newblock What do people value when they negotiate?
\newblock \emph{Journal of Personality and Social Psychology}, 91(3):493--512.

\bibitem[{Dong et~al.(2023)Dong, Jiang, Jin, and Li}]{dong2023self}
Yihong Dong, Xue Jiang, Zhi Jin, and Ge~Li. 2023.
\newblock Self-collaboration code generation via chatgpt.
\newblock \emph{arXiv preprint arXiv:2304.07590}.

\bibitem[{Fisher et~al.(1991)Fisher, Ury, and Patton}]{fisher1991}
Roger Fisher, William~L. Ury, and Bruce Patton. 1991.
\newblock \emph{Getting to Yes: Negotiating Agreement Without Giving In}, 2
  edition.
\newblock Penguin Books.

\bibitem[{Fitzpatrick et~al.(2017)Fitzpatrick, Darcy, and
  Vierhile}]{fitzpatrick2017}
Kathleen~K. Fitzpatrick, Alison Darcy, and Molly Vierhile. 2017.
\newblock Delivering cbt using a fully automated conversational agent.
\newblock \emph{JMIR Mental Health}, 4(2):e19.

\bibitem[{Guo et~al.(2024)Guo, Chen, Wang, Chang, Pei, Chawla, Wiest, and
  Zhang}]{guo2024large}
Taicheng Guo, Xiuying Chen, Yaqi Wang, Ruidi Chang, Shichao Pei, Nitesh~V
  Chawla, Olaf Wiest, and Xiangliang Zhang. 2024.
\newblock Large language model based multi-agents: A survey of progress and
  challenges.
\newblock \emph{Proc. of IJCAI}.

\bibitem[{Han et~al.(2024)Han, Zhang, Yao, Jin, Xu, and He}]{han2024llm}
Shanshan Han, Qifan Zhang, Yuhang Yao, Weizhao Jin, Zhaozhuo Xu, and Chaoyang
  He. 2024.
\newblock Llm multi-agent systems: Challenges and open problems.
\newblock \emph{arXiv preprint arXiv:2402.03578}.

\bibitem[{Hong et~al.(2023)Hong, Zheng, Chen, Cheng, Wang, Zhang, Wang, Yau,
  Lin, Zhou et~al.}]{hong2023metagpt}
Sirui Hong, Xiawu Zheng, Jonathan Chen, Yuheng Cheng, Jinlin Wang, Ceyao Zhang,
  Zili Wang, Steven Ka~Shing Yau, Zijuan Lin, Liyang Zhou, and 1 others. 2023.
\newblock Metagpt: Meta programming for multi-agent collaborative framework.
\newblock \emph{arXiv preprint arXiv:2308.00352}.

\bibitem[{Hua et~al.(2024)Hua, Qu, and Haffari}]{hua2024assistive}
Yuncheng Hua, Lizhen Qu, and Gholamreza Haffari. 2024.
\newblock Assistive large language model agents for socially-aware negotiation
  dialogues.
\newblock \emph{arXiv preprint arXiv:2402.01737}.

\bibitem[{Kamoi et~al.(2024)Kamoi, Zhang, Zhang, Han, and Zhang}]{kamoi2024}
Ryo Kamoi, Yusen Zhang, Nan Zhang, Jiawei Han, and Rui Zhang. 2024.
\newblock When can {LLMs} actually correct their own mistakes? {A} critical
  survey of self-correction of {LLMs}.
\newblock \emph{arXiv preprint arXiv:2406.01297}.

\bibitem[{Lande(2022)}]{lande2022}
John Lande. 2022.
\newblock The critical importance of pre-session preparation in mediation.
\newblock University of Missouri School of Law.

\bibitem[{Li et~al.(2024)Li, Wen, Wang, Li, Yuan, Liu, Liu, Xu, Wang, Sun
  et~al.}]{li2024personal}
Yuanchun Li, Hao Wen, Weijun Wang, Xiangyu Li, Yizhen Yuan, Guohong Liu,
  Jiacheng Liu, Wenxing Xu, Xiang Wang, Yi~Sun, and 1 others. 2024.
\newblock Personal llm agents: Insights and survey about the capability,
  efficiency and security.
\newblock \emph{arXiv preprint arXiv:2401.05459}.

\bibitem[{Liang et~al.(2023)}]{liang2023}
T.~Liang and 1 others. 2023.
\newblock \href {https://arxiv.org/abs/2305.19118} {Encouraging divergent
  thinking through multi-agent debate}.
\newblock \emph{arXiv preprint}.

\bibitem[{Lin et~al.(2014)Lin, Gal, Kraus, and Mazliah}]{lin2014}
Raz Lin, Ya'akov Gal, Sarit Kraus, and Yair Mazliah. 2014.
\newblock Training with automated agents improves people's behavior in
  negotiation and coordination tasks.
\newblock \emph{Decision Support Systems}, 60:1--9.

\bibitem[{Madaan et~al.(2023)}]{madaan2023}
Aman Madaan and 1 others. 2023.
\newblock Self-refine: Iterative refinement with self-feedback.
\newblock In \emph{Advances in Neural Information Processing Systems},
  volume~36.

\bibitem[{McAleese et~al.(2024)}]{mcaleese2024}
N.~McAleese and 1 others. 2024.
\newblock \href {https://arxiv.org/abs/2407.00215} {Llm critics help catch llm
  bugs}.
\newblock \emph{arXiv preprint}.

\bibitem[{Moore(2014)}]{moore2014}
Christopher~W. Moore. 2014.
\newblock \emph{The Mediation Process}, 4 edition.
\newblock Jossey-Bass.

\bibitem[{Mushtaq et~al.(2025)Mushtaq, Naeem, Ghaznavi, Taj, Hashmi, and
  Qadir}]{mushtaq2025harnessing}
Abdullah Mushtaq, Rafay Naeem, Ibrahim Ghaznavi, Imran Taj, Imran Hashmi, and
  Junaid Qadir. 2025.
\newblock Harnessing multi-agent llms for complex engineering problem-solving:
  A framework for senior design projects.
\newblock In \emph{2025 IEEE Global Engineering Education Conference (EDUCON)},
  pages 1--10. IEEE.

\bibitem[{Olekalns and Adair(2014)}]{olekalns2014}
Mara Olekalns and Wendi~L. Adair. 2014.
\newblock \emph{Handbook of Research on Negotiation}.
\newblock Edward Elgar Publishing.

\bibitem[{Passmore et~al.(2025)Passmore, Olafsson, and Tee}]{passmore2025}
Jonathan Passmore, B.~Olafsson, and D.~Tee. 2025.
\newblock A systematic literature review of ai in coaching.
\newblock \emph{Journal of Work-Applied Management}.

\bibitem[{Qian et~al.(2023)Qian, Cong, Yang, Chen, Su, Xu, Liu, and
  Sun}]{qian2023communicative}
Chen Qian, Xin Cong, Cheng Yang, Weize Chen, Yusheng Su, Juyuan Xu, Zhiyuan
  Liu, and Maosong Sun. 2023.
\newblock Communicative agents for software development.
\newblock \emph{arXiv preprint arXiv:2307.07924}, 6.

\bibitem[{Rasal(2024)}]{rasal2024llm}
Sumedh Rasal. 2024.
\newblock Llm harmony: Multi-agent communication for problem solving.
\newblock \emph{arXiv preprint arXiv:2401.01312}.

\bibitem[{Sharma et~al.(2023)}]{sharma2023}
M.~Sharma and 1 others. 2023.
\newblock \href {https://arxiv.org/abs/2310.13548} {Towards understanding
  sycophancy in language models}.
\newblock \emph{arXiv preprint}.

\bibitem[{Sun et~al.(2025)Sun, Li, Dong, Liu, Xu, Li, and Liu}]{sun2025multi}
Songtao Sun, Jingyi Li, Yuanfei Dong, Haoguang Liu, Chenxin Xu, Fuyang Li, and
  Qiang Liu. 2025.
\newblock Multi-agent application system in office collaboration scenarios.
\newblock \emph{arXiv preprint arXiv:2503.19584}.

\bibitem[{Tan et~al.(2024)Tan, Westermann, Pottanigari, {\v{S}}avelka,
  Mee{\`u}s, Godet, and Benyekhlef}]{tan2024robots}
Jinzhe Tan, Hannes Westermann, Nikhil~Reddy Pottanigari, Jarom{\'\i}r
  {\v{S}}avelka, S{\'e}bastien Mee{\`u}s, Mia Godet, and Karim Benyekhlef.
  2024.
\newblock Robots in the middle: Evaluating llms in dispute resolution.
\newblock In \emph{Legal Knowledge and Information Systems}, pages 168--179.
  IOS Press.

\bibitem[{Tao et~al.(2024)Tao, Zhou, Zhang, and Cheng}]{tao2024magis}
Wei Tao, Yucheng Zhou, Wenqiang Zhang, and Yu~Cheng. 2024.
\newblock Magis: Llm-based multi-agent framework for github issue resolution.
\newblock \emph{arXiv preprint arXiv:2403.17927}.

\bibitem[{Tessler et~al.(2024)Tessler, Bakker, Jarrett, Sheahan, Chadwick,
  Koster, Evans, Campbell-Gillingham, Collins, Parkes et~al.}]{tessler2024ai}
Michael~Henry Tessler, Michiel~A Bakker, Daniel Jarrett, Hannah Sheahan,
  Martin~J Chadwick, Raphael Koster, Georgina Evans, Lucy Campbell-Gillingham,
  Tantum Collins, David~C Parkes, and 1 others. 2024.
\newblock Ai can help humans find common ground in democratic deliberation.
\newblock \emph{Science}, 386(6719):eadq2852.

\bibitem[{Thompson(2010)}]{thompson2010}
Leigh~L. Thompson. 2010.
\newblock \emph{The Mind and Heart of the Negotiator}, 5 edition.
\newblock Pearson.

\bibitem[{Tran et~al.(2025)Tran, Dao, Nguyen, Pham, O'Sullivan, and
  Nguyen}]{tran2025multi}
Khanh-Tung Tran, Dung Dao, Minh-Duong Nguyen, Quoc-Viet Pham, Barry O'Sullivan,
  and Hoang~D Nguyen. 2025.
\newblock Multi-agent collaboration mechanisms: A survey of llms.
\newblock \emph{arXiv preprint arXiv:2501.06322}.

\bibitem[{Triantafyllopoulos and
  Kalles(2025)}]{triantafyllopoulos2025divergence}
Loukas Triantafyllopoulos and Dimitris Kalles. 2025.
\newblock From divergence to alignment: Evaluating the role of large language
  models in facilitating agreement through adaptive strategies.
\newblock \emph{Future Internet}, 17(9):407.

\bibitem[{Wei et~al.(2024)}]{wei2024}
Jason Wei and 1 others. 2024.
\newblock \href {https://arxiv.org/abs/2308.03958} {Simple synthetic data
  reduces sycophancy in large language models}.
\newblock \emph{arXiv preprint}.

\bibitem[{Westermann et~al.(2023)Westermann, Savelka, and
  Benyekhlef}]{westermann2023llmediator}
Hannes Westermann, Jaromir Savelka, and Karim Benyekhlef. 2023.
\newblock Llmediator: Gpt-4 assisted online dispute resolution.
\newblock \emph{arXiv preprint arXiv:2307.16732}.

\bibitem[{Wissler and Hinshaw(2022)}]{wissler2022}
Roselle~L. Wissler and A.~Hinshaw. 2022.
\newblock What happens before the first mediation session? an empirical study
  of pre-session communications.
\newblock \emph{Cardozo Journal of Conflict Resolution}, 23(1):143--198.

\bibitem[{Zeng et~al.(2024)Zeng, Wu, Zhang, Wang, and Wu}]{zeng2024autodefense}
Yifan Zeng, Yiran Wu, Xiao Zhang, Huazheng Wang, and Qingyun Wu. 2024.
\newblock Autodefense: Multi-agent llm defense against jailbreak attacks.
\newblock \emph{arXiv preprint arXiv:2403.04783}.

\bibitem[{Zhang et~al.(2025{\natexlab{a}})Zhang, Cui, Wang, Zhang, Wang, Wu,
  and Hu}]{zhang2025if}
Hangfan Zhang, Zhiyao Cui, Xinrun Wang, Qiaosheng Zhang, Zhen Wang, Dinghao Wu,
  and Shuyue Hu. 2025{\natexlab{a}}.
\newblock If multi-agent debate is the answer, what is the question.
\newblock \emph{arXiv preprint arXiv:2502.08788}.

\bibitem[{Zhang et~al.(2025{\natexlab{b}})Zhang, Chen, Yeh, and
  Li}]{zhang2025metamind}
Xuanming Zhang, Yuxuan Chen, Min-Hsuan Yeh, and Yixuan Li. 2025{\natexlab{b}}.
\newblock Metamind: Modeling human social thoughts with metacognitive
  multi-agent systems.
\newblock \emph{arXiv preprint arXiv:2505.18943}.

\bibitem[{Zhou et~al.(2019)Zhou, He, Black, and Tsvetkov}]{zhou2019dynamic}
Yiheng Zhou, He~He, Alan~W Black, and Yulia Tsvetkov. 2019.
\newblock A dynamic strategy coach for effective negotiation.
\newblock \emph{arXiv preprint arXiv:1909.13426}.

\bibitem[{Zhu et~al.(2024)Zhu, Zhang, Huang, Li, Niu, Fan, Lun, Tao, Su, Gong
  et~al.}]{zhu2024plangpt}
He~Zhu, Wenjia Zhang, Nuoxian Huang, Boyang Li, Luyao Niu, Zipei Fan, Tianle
  Lun, Yicheng Tao, Junyou Su, Zhaoya Gong, and 1 others. 2024.
\newblock Plangpt: Enhancing urban planning with tailored language model and
  efficient retrieval.
\newblock \emph{arXiv preprint arXiv:2402.19273}.

\end{thebibliography}
\appendix
\section{Prompt Templates}
\label{sec:prompts}
This appendix provides the complete system and user prompts for each LLM module described in Section~\ref{sec:architecture}. We refer to the Study~1 (more affirming) and Study~2 (refined, less affirming) variations of the dialogue and critic prompts as the \emph{Affirming} and \emph{Neutral} variants. Lines marked \texttt{[NEUTRAL ONLY]} are present in the Study~2 prompts but not in Study~1.
\subsection{Pre-Mediation Dialogue Agent}
\label{app:prompt-dialogue}
The primary conversational module, guiding users through the eight-phase preparation protocol.
\begin{lstlisting}
You are an expert mediator conducting a one-on-one coaching session with an individual preparing for a negotiation. Your goal is to effectively guide the person through a structured, empathetic, and constructive preparation process to improve their readiness for the mediation. Your role is strictly to ask questions that help the individual articulate and clarify their own preferences, priorities, and boundaries. You should avoid suggesting specific outcomes yourself.
[NEUTRAL ONLY] You should limit affirming the user's beliefs excessively; only affirm if responses are genuinely helpful or constructive for the upcoming negotiation.
When you send a message, it will first be sent to a critic agent. The critic agent will reply with either APPROVED, WARNING, or REJECTED for your message being sent to the user. If you receive a WARNING or REJECTED, use the reasoning it provides to modify your message accordingly.
The conversation is structured around the following phases:
Phase 1: Rapport Building & Introduction
 - Welcome the participant warmly. Introduce yourself naturally as their mediator.
 - Do not mention being an AI or technology in your introduction.
 - If directly asked, "Are you an AI?" Acknowledge honestly.
Phase 2: In-Depth Exploration of Preferences
 - Explore their preferences, ideal arrangements, importance, boundaries, past experiences, success definition
Phase 3: Prioritization and Trade-offs
 - Help identify the most important topics vs. flexible areas
Phase 4: Perspective-Taking
 - Invite consideration of roommates' views and needs
Phase 5: Emotional Awareness and Conflict Management
 - Explore how the participant handles disagreements
Phase 6: Confidence and Preparedness
 - Ask about the sense of preparedness for the upcoming negotiation
Phase 7: Principles & Relationship Maintenance
 - Reflect on the relationship with the other parties
Phase 8: Closing and Wrap-Up
 - Affirm progress, offer reassurance, invite final questions
GENERAL STYLE & ABSOLUTE RULES:
 - CRITICAL: Keep responses SHORT like texting (1 to 3 sentences)
 - Avoid bullet points, numbered lists, or formatted structure
 - Generally ask one question at a time
 - Never use em dashes or en dashes
 - Do not offer solutions
 [NEUTRAL ONLY] Balance warmth and affirmation with friction or pushing back on ideas when productive
\end{lstlisting}
\subsection{Critic Agent}
\label{app:prompt-critic}
The oversight module that reviews each candidate response before it is sent to the user. The Neutral variant adds a WARNING tier and additional rejection criteria.
\begin{lstlisting}
You are the Critic Agent assigned to supervise and refine the behavior of a Mediator Agent in a conversational setting. Your role is to ensure that every message the mediator sends to the user is appropriate, engaging, logically structured, and aligned with conversational best practices.
You operate behind the scenes. The user will never see your messages.
WORKFLOW INSTRUCTIONS
Message Approval (STRICT OUTPUT FORMAT):
  - APPROVED
  [NEUTRAL ONLY] - WARNING: [your suggestion here]
  - REJECTED: [your reason here]
APPROVAL CRITERIA
 - Non-repetitive, maintains logical flow
 - Substantively advances the conversation
 - Challenges productively (not purely validating)
 - Appropriate tone, ethical boundaries
 - Clear and understandable
[NEUTRAL ONLY] WARNING CRITERIA (minor issues)
 - Could include stronger probing question
 - Slightly repetitive in style
 - Smoother topic transition needed
 - Does not advance the conversation
REJECTION CRITERIA (REJECT if any apply)
 - MESSAGE TOO LONG: Exceeds 3 sentences or ~320 characters
 - TOO MANY QUESTIONS: More than one question mark
 - Repeats question/phrase from earlier conversation
 - Uses bullet points, numbered lists
 - Contains em dashes or en dashes
 [NEUTRAL ONLY] Additional rejection criteria:
   - Purely validating without substance
   - Accepts vague answer without probing
TOPIC COVERAGE MONITORING
Track: Chores, Guests, Quiet Hours
 - WARNING if one topic lightly covered before closing
 - REJECTION if topic is completely skipped
MONITORING TASKS
 - User Boredom
 - Substance Check
 - Repetition Check
 [NEUTRAL ONLY] Excessive Affirmation Check
\end{lstlisting}
\subsection{User Prediction Agent}
\label{app:prompt-prediction}
Analyzes user messages and produces structured JSON predictions across SVI-derived dimensions with rationales.
\begin{lstlisting}
SYSTEM PROMPT:
You are an AI that analyzes negotiation messages and provides structured assessments.
USER PROMPT:
Analyze this message in a roommate negotiation context and evaluate the following parameters:
USER'S LATEST MESSAGE: "${messageContent}"
PREVIOUS USER MESSAGES: ${JSON.stringify(userMessages)}
MEDIATOR MESSAGES: ${JSON.stringify(mediatorMessages)}
Evaluate on these dimensions (1-5 scale):
1. chore_schedule_importance
2. guests_frequency_importance
3. quiet_hours_importance
4. frustration_expression
5. trust_mediator
6. outcome_confidence
7. balanced_outcome_preparedness
8. negotiation_confidence
9. principles_preparedness
10. perspective_understanding
11. relationship_importance
For each evaluation, provide a brief rationale explaining your reasoning.
Return the analysis in JSON format with a 'rationale' object.
\end{lstlisting}
\subsection{Summary Generation Agent}
\label{app:prompt-summary}
Synthesizes the full transcript and prediction output into a one-page report intended for the human mediator.
\begin{lstlisting}
SYSTEM PROMPT:
You are an expert mediator who creates concise, insightful analysis reports for human mediators based on roommate negotiation conversations.
USER PROMPT:
You are an expert mediator analyzing a pre-mediation session. Your task is to create a concise report (no more than one page) for a human mediator who will assist these roommates. Use the full transcript of the conversation and the results from the prediction agent in your report.
Please create a focused, one-page report that includes:
1. PARTICIPANT PROFILE
   A brief summary of the individual's key characteristics and negotiation style.
2. PRIORITIZED PREFERENCES
   - Rank their preferences regarding chore schedules, guest frequency, and quiet hours based on importance
   - Highlight their non-negotiable points versus areas where they are flexible
3. COMMUNICATION STYLE
   - How they express frustration or disagreement
   - Their level of trust in the mediation process
4. MEDIATION APPROACH RECOMMENDATIONS
   - Suggested conversation starters tailored to this individual
   - Topics that might require special handling due to sensitivity
   - Potential compromise areas the mediator should explore
Format the report in plain text with clear headings. Keep it concise - it should not exceed one page when printed.
\end{lstlisting}
\paragraph{Model and decoding parameters.} All LLM modules use GPT-4o (accessed via the OpenAI Chat Completions API). The optional voice-to-text module uses Whisper-1. We used default API parameters; specific model snapshot versions, temperature, and \texttt{max\_tokens} values are recorded in the released codebase upon publication.
\end{document}